\documentclass[10pt,twocolumn]{article}

\usepackage[a4paper,margin=0.75in]{geometry}

\usepackage[T1]{fontenc}
\usepackage[utf8]{inputenc}
\usepackage{times}
\usepackage{microtype}

\usepackage[hidelinks]{hyperref}
\usepackage{url}
\usepackage[small]{caption}
\usepackage{graphicx}
\usepackage{booktabs}
\usepackage{xcolor}
\usepackage{soul}

\usepackage{amsmath}
\usepackage{amssymb}
\usepackage{amsthm}

\usepackage{algorithm}
\usepackage{algorithmic}

\usepackage{listings}
\usepackage{lstautogobble}

\usepackage{natbib}
\setcitestyle{authoryear,round,maxcitenames=2}

\newtheorem{example}{Example}

\title{
    \Large\bfseries
    DeepLog: A Software Framework for Modular Neurosymbolic AI
}

\author{
\begin{tabular}{c}
Robin Manhaeve, Stefano Colamonaco, Vincent Derkinderen, \\
Rik Adriaensen, Lucas Van Praet, Luc De Raedt, Giuseppe Marra \\
\\
Department of Computer Science and Leuven.AI \\
KU Leuven, Belgium \\
\texttt{\{firstname.lastname\}@kuleuven.be}
\end{tabular}
}

\date{}

\begin{document}

\maketitle


\begin{abstract} 
\emph{DeepLog} is an operational neurosymbolic framework that unifies logic and deep learning within standard PyTorch workflows. 
While existing neurosymbolic systems focus on a particular paradigm and semantics, DeepLog serves as a universal backend that can emulate many systems in the neurosymbolic ``alphabet soup''.
By treating diverse neurosymbolic languages as high-level specifications, the DeepLog software automatically compiles them into optimized arithmetic circuits. This design lowers the barrier for machine learning practitioners by treating logic as composable modules, while providing neurosymbolic developers with a shared, high-performance basis for prototyping new integration strategies.
The code is available here: \url{https://github.com/ML-KULeuven/deeplog}
\end{abstract}


\section{Introduction}

Neurosymbolic AI (NeSy) aims to combine  
deep learning with 
symbolic reasoning and representations \cite{hitzler2022neuro,garcez2023neurosymbolic,MARRA2024104062}. 
However, the current landscape resembles an ``alphabet soup'' of seemingly disconnected paradigms
and frameworks. Several dimensions are commonly used to distinguish among these approaches, including the nature of their semantics (e.g., fuzzy or probabilistic), the integration paradigm (e.g., incorporation of symbolic representations into the loss function or the architecture of the machine learning model), and the type of symbolic front-end employed (e.g., first-order logic or logic programming).
Because these differences are often treated as fundamental rather than alternative instantiations, new approaches are generally pursued within non-interoperable frameworks. As a result, no cross-paradigm software tool has yet emerged, slowing systematic comparison and cumulative progress in the field.

To address this gap, we contribute and demo the DeepLog software that provides building blocks and primitives for neurosymbolic AI, enabling abstraction over commonly used representations and computational mechanisms. It is based on the DeepLog abstract neurosymbolic machine defined in ~\cite{derkinderen2025deeplog}, and can be used to easily implement, reconstruct and compare a wide range of neurosymbolic systems. To this end, it consolidates essential components from the existing “alphabet soup” of approaches and offers the necessary primitives for designing new ones. 

The DeepLog software exploits modern deep learning soft- and hardware.  First, it 
 extends standard PyTorch modules with symbolic annotations on their inputs and outputs, 
 based on formulas expressed in the DeepLog language~\cite{derkinderen2025deeplog},  enabling the software to manage composition and integration automatically. 
 Second, it does not only use GPU-acceleration for neural modules, but also for its symbolic components.  
Most other neurosymbolic AI systems implement their symbolic component on the CPU, which not only degrades performance due to data transfer and synchronization, but also introduces additional complications for debugging, and potential errors in terms of what data is on what device.

\begin{figure}
    \centering
    \includegraphics[width=1.0\linewidth]{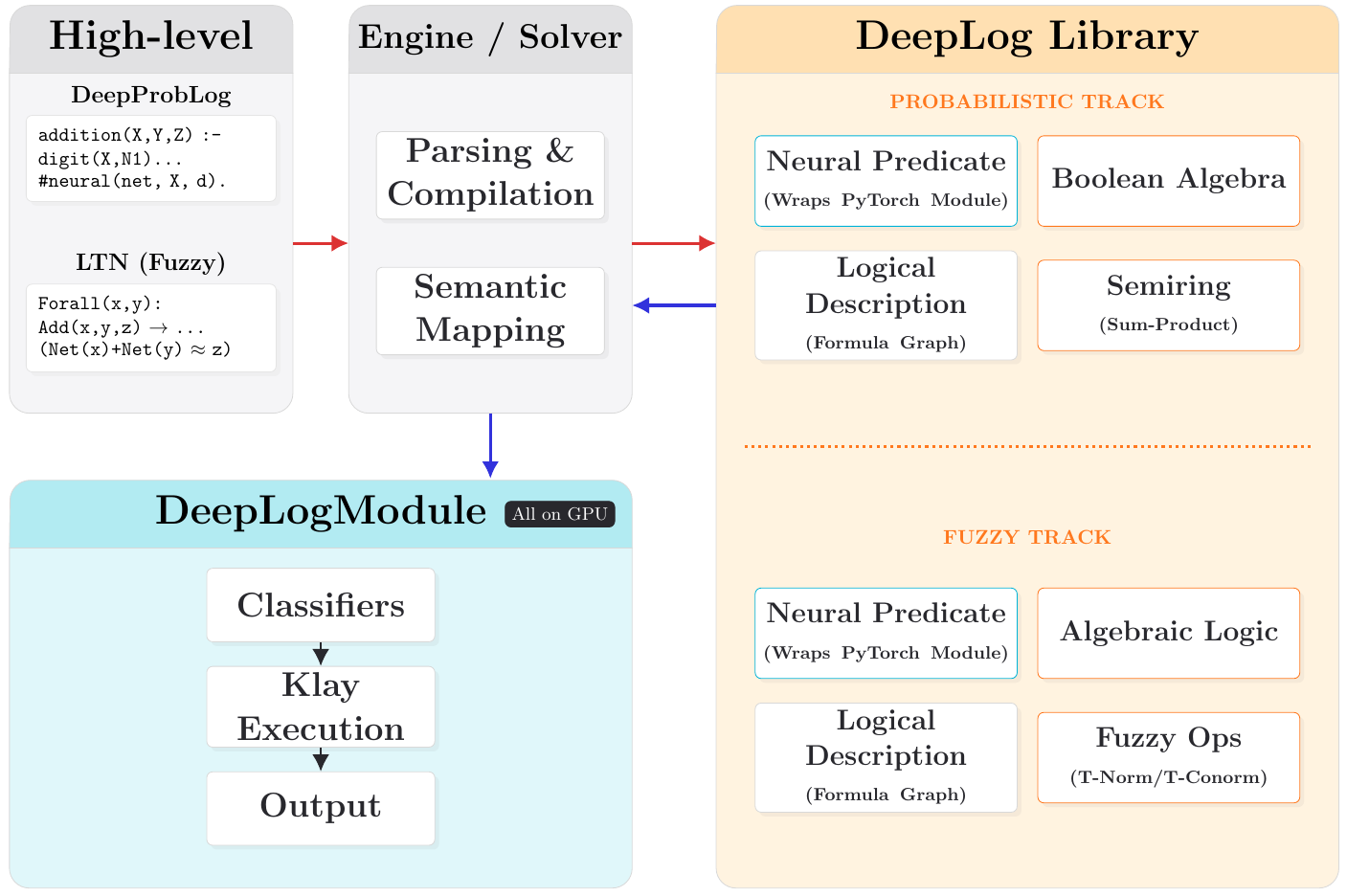}
    \caption{The High-Level Specification (top-left) allows users to switch between semantics simply by changing the underlying logic track in the DeepLog Library (right). The system automatically compiles these specifications into an optimized DeepLogModule (bottom-left) for execution on the GPU.}
    \label{fig:visabstract}
\end{figure}

While most existing NeSy systems adopt a monolithic design encompassing model specification, data handling, and training, DeepLog is designed to be modular and non-intrusive. Rather than replacing established machine learning workflows, it can be integrated selectively into existing pipelines. This design choice lowers the barrier to entry for machine learning practitioners interested in neurosymbolic AI, while simultaneously providing framework developers with a flexible basis for prototyping new ideas without sacrificing performance.

\noindent \textbf{The main contribution} of this work is an implementation of DeepLog, an abstract neurosymbolic machine, that
\begin{itemize}
    \itemsep0em
    \item allows us to reconstruct any member of the alphabet soup with minimal code, 
    \item provides out-of-the-box GPU acceleration, 
    \item provides modularity as automatic module compositions, 
    \item integrates with modern ML software (PyTorch).
\end{itemize}


\section{Bridging Paradigms}

The integration of neurosymbolic systems remains largely an open challenge, with only a few approaches explicitly aiming at unification.  Most of them focus on the uniform encoding of multiple types of semantics, such as LYRICS \cite{marra2019lyrics} and Pylon \cite{ahmed2022pylon}. However, these frameworks are usually limited to a specific neurosymbolic paradigm, most often semantic-based regularization, where logical constraints are added as loss terms during training. In addition, they rely on a single front-end representation, which limits their use across the broader neurosymbolic landscape.
More recent work, such as ULLER \cite{van2024uller}, attempts to overcome this limitation by introducing a unified language that can express a wide range of neurosymbolic specifications. However, this unification mainly takes place at the high-level front-end layer. At present, no public implementation is available, so its practical usefulness is still unclear. Moreover, because the unification is restricted to the front-end, it has a strong impact on user interaction, as different research communities prioritize different syntactic structures and modeling abstractions.

\section{DeepLog for ML Practitioners}

This section introduces DeepLog from the perspective of machine learning practitioners who wish to integrate symbolic reasoning into existing PyTorch-based workflows. The central idea is that symbolic information attached to module interfaces enables automatic composition, validation, and transformation of neurosymbolic components, while remaining compatible with standard deep learning practices.

\subsection{Module Composition}

Neurosymbolic models are typically constructed by composing heterogeneous components, such as neural network predictors, logical functions or constraints. Each of these components assumes that its inputs represent specific semantic quantities, for example Boolean values, probabilities, or continuous fuzzy scores. In practice, these assumptions are rarely made explicit, leading to fragile compositions and error-prone glue code.
DeepLog addresses this through \texttt{DeepLogModules}, which are standard PyTorch modules augmented with symbolic annotations on their inputs and outputs. These annotations specify the semantic role of each tensor and are represented using (tuples of) \texttt{SymTensors}: N-dimensional tensors whose entries are associated with symbolic expressions. 

These symbolic interfaces allow DeepLog to automatically compose modules. When chaining two modules, DeepLog automatically derives and inserts the necessary GPU-accelerated transformations between them. The annotations also enable run-time validation of module interfaces, ensuring that inputs and outputs satisfy the expected symbolic structure, significantly improving debuggability.

DeepLog supports two primary composition patterns. First, modules can be composed sequentially, analogous to \texttt{torch.nn.Sequential}, where the output of one module feeds directly into the next. Second, when a collection of modules forms a directed acyclic graph based on their symbolic input--output dependencies, DeepLog can automatically wire the full computation graph, including all required intermediate transformations. By leveraging PyTorch’s asynchronous execution model, this approach maximizes parallelism and GPU utilization.

\begin{example}[Semantic loss from a DIMACS constraint]
We demonstrate semantic loss~\cite{xu2018semantic} using a logical constraint loaded from a DIMACS CNF specification.
Assume a classifier produces probabilities for classes \(\{A, B, C\}\), and domain knowledge enforces a hierarchy:
\(A \Rightarrow B\) and \(C \Rightarrow B\).
This constraint is encoded once as a Boolean CNF and reused independently of the neural model.

\begin{lstlisting}[language=Python, breaklines=true, basicstyle=\ttfamily\small, columns=fullflexible, autogobble=true, showstringspaces=false]
dimacs = """
p cnf 3 2
-1 2 0   c A -> B
-3 2 0   c C -> B
"""

phi = parse_dimacs_to_module(dimacs, structure='probability')
expected = SymTensor(["A_p","B_p","C_p"])
phi = reshape_input(phi, expected)
\end{lstlisting}
Phi is a PyTorch module that now fits into a normal training loop:
\begin{lstlisting}[language=Python, breaklines=true, basicstyle=\ttfamily\small, columns=fullflexible, autogobble=true]
...
semantic_loss = -(phi(y_pred).log()).mean()
loss = classification_loss + semantic_loss
\end{lstlisting}
\end{example}

\subsection{GPU-Accelerated Circuits}

A second key component of DeepLog is its use of GPU-accelerated circuits for evaluating Boolean formulas, probabilistic expressions, and related symbolic computations. These circuits provide an efficient inference structure for reasoning tasks that would otherwise be executed on the CPU, creating a performance bottleneck in hybrid systems. In DeepLog, we rely on the KLay~\cite{klay} package for efficient circuit evaluation; comparative performance results are summarized in Table \ref{tab:eval_time}.
To enforce structural properties such as determinism or decomposability, DeepLog targets specialized circuit backends, including representations based on sentential decision diagrams~\cite{darwiche2011sdd}. 

\begin{table}[h]
    \centering
    \setlength{\tabcolsep}{5pt} 
    \begin{tabular}{cccc}
        \toprule
         & \multicolumn{3}{c}{Inference time [s/query]} \\
        \cmidrule(r){2-4}
        Digits & DPL CPU & DeepLog CPU & DeepLog GPU \\
        \midrule
        1 & $6.14 \times 10^{-3}$ & $1.09 \times 10^{-5}$ & $5.62 \times 10^{-7}$ \\
        2 & $3.52 \times 10^{-2}$ & $4.13 \times 10^{-5}$ & $1.52 \times 10^{-6}$ \\
        3 & $1.32 \times 10^{-1}$ & $2.43 \times 10^{-3}$ & $1.52 \times 10^{-4} $ \\
        4 & -- & $1.96 \times 10^{-1}$ & $5.40 \times 10^{-2}$ \\
        \bottomrule
    \end{tabular}
    \caption{From \protect\cite{derkinderen2025deeplog}, the average time in seconds to evaluate the neural network and arithmetic circuit for one query in the MNIST-Addition task. DeepProbLog (DPL) is included as a reference. DeepLog with GPU acceleration is most performant.}
    \label{tab:eval_time}
    
\end{table}

\section{DeepLog as a Backend Architecture for NeSy Systems}

Different NeSy systems appear fundamentally different, yet in practice they all execute the same pipeline: enumerate symbolic structures, attach neural scores, and aggregate them. This duplication obscures real design trade-offs and leads to CPU-bound, hard-to-optimize implementations.
At a semantic level, inference in virtually all NeSy systems reduces to evaluating a single formula~\cite{de2025defining}. Apparent differences arise from how this formula is calculated, not from what is computed.
The DeepLog abstract machine~\cite{derkinderen2025deeplog} was introduced to make this computation explicit. It serves as a semantic intermediate representation and abstract machine target for NeSy systems, separating front-end symbolic languages from back-end execution strategies. 

Languages such as DeepProbLog~\cite{manhaeve2018deepproblog} and NeurASP~\cite{yang2023neurasp} use a Prolog- and ASP-like language, respectively. The base languages are lifted to the NeSy level through inclusion of the neural predicate, which parameterizes the probability of each model of the logic. Such a system would be implemented by first gathering proofs from a solver or engine. These proofs, whose symbolic information is backed by neural networks that provide their probabilities, are then used to calculate the probability of the query. While this process can be optimized (cf. knowledge compilation), without careful design the inference is typically still CPU-bound.

In DeepLog, the process is closer to Figure \ref{fig:visabstract}. The solver or engine does not construct a proof, but instead calls methods from a factory object in the DeepLog library, which immediately returns an instantiation of a DeepLog module. 
This factory can be configured to rely on certain components, or on techniques of choice (e.g. circuits versus sampling based approaches). 
\begin{example}
To demonstrate DeepLog's support in developing neurosymbolic systems, we show how one could for example implement logic tensor network~(LTN)~\cite{badreddine2022logic}. 
LTN is a neuro-symbolic system where fuzzy predicate logic formulas are transformed into Tensorflow modules. In it, entities are represented with tensors, predicates are functions that map tensors onto fuzzy scores, and quantification is implemented using aggregation.
Here we give the definitions of the separate components:
\begin{lstlisting}[language=Python, breaklines=true, basicstyle=\ttfamily\small, columns=fullflexible, autogobble=true, showstringspaces=false]
ltn_fuzzy = {
    'not': lambda x: 1 - x,
    'and': lambda x, y: x * y,
    ...
}
exists = AggregationModule(name='exists',
                op=lambda x: p_mean(x, 6)
...
class EqualityPredicate(Predicate):
    functor, arity = 'eq', 2
    structure = 'fuzzy'

    def forward_predicate(self, x, y):
        return (-(x - y).norm(dim=1)).exp()
\end{lstlisting}

These can now be combined into a single factory object, with a one-to-one mapping from LTN concepts to factory methods:
\begin{lstlisting}[language=Python, breaklines=true, basicstyle=\ttfamily\small, columns=fullflexible, autogobble=true, showstringspaces=false]
ltn = DeepLogModuleFactory(
    structures = {'fuzzy': ltn_fuzzy},
    aggregators = {'exists': exists, ...},
    ... 
    predicates = [EqualityPredicate],
)

Not = lambda x: ltn.unary_node('not', x)
Or = lambda x,y: ltn.binary_node('pr', x, y)
...
\end{lstlisting}

\end{example}

This configurability of the DeepLog backend allows us to easily digest the NeSy alphabet soup, as we can easily substitute probabilistic logic with fuzzy logic, and more generally compare inference techniques while keeping the remainder fixed.


\section{Conclusion and future work}

We introduced \emph{DeepLog}, an operational framework that serves as a neurosymbolic abstract machine bridging symbolic reasoning and deep learning within standard PyTorch workflows. By separating semantic specification from operational realization, DeepLog enables automatic composition, validation, and efficient execution of neurosymbolic models on modern accelerators. Through symbolically annotated modules, a semantic intermediate language, and a compiler-style factory, the framework supports both ML practitioners seeking lightweight integration of symbolic constraints and NeSy developers building new languages and inference paradigms on a shared backend.

Several directions for future work remain. A primary focus is the automation of more aggressive circuit-level optimizations. While DeepLog already supports multiple compilation strategies and circuit backends, we aim to incorporate systematic optimization techniques~\cite{atlas}, including structure-aware rewritings, fusion of aggregation and inference, and others. Automating such transformations would further reduce manual engineering effort while improving performance and scalability.
Beyond circuit optimization, future work includes providing alternative inference mechanisms such as sampling.


\section*{Acknowledgements}
This work project has received funding from the European Research Council (ERC) under the European Union’s Horizon 2020 research and innovation programme (Grant agreement No. 101142702).
This research received funding from the Flemish Government under the “Onderzoeksprogramma Artificiële Intelligentie (AI) Vlaanderen” programme. 

\bibliographystyle{abbrvnat}
\bibliography{bibliography}

\end{document}